\documentclass[11pt,a4paper]{article}
\usepackage[hyperref]{emnlp-ijcnlp-2019}
\usepackage{times}
\usepackage{latexsym}
\usepackage{multirow}
\usepackage{graphicx}
\usepackage{siunitx}
\usepackage{booktabs}
\usepackage{amsmath}
\usepackage{amsfonts}
\usepackage{enumitem}
\usepackage{url}
\usepackage{color}
\usepackage{url}

\aclfinalcopy % Uncomment this line for the final submission

\title{Do Multi-hop Readers Dream of Reasoning Chains?}

\author{Haoyu Wang \thanks{\, Equal contributions.} $^{\dagger}$ \quad \quad Mo Yu  $^{*\dagger}$ \quad \quad Xiaoxiao Guo  $^{*\dagger}$ \quad \quad \textbf{Rajarshi Das} $^{*\ddagger}$ \\ \textbf{Wenhan Xiong} $^{*\S}$ \quad \quad \textbf{Tian Gao} $^{*\dagger}$\\
$^\dagger$ IBM Research \quad \quad $^\ddagger$ Umass Amherst \quad \quad $^\S$ UC Santa Barbara 
}

\date{}

\begin{document}

\maketitle

\begin{abstract}
General Question Answering (QA) systems over texts require the multi-hop reasoning capability, i.e. the ability to reason with information collected from multiple passages to derive the answer. 
In this paper we conduct a systematic analysis to assess such an ability of various existing models proposed for multi-hop QA tasks.
Specifically, our analysis investigates that  whether providing the full reasoning chain of multiple passages, instead of just one final passage where the answer appears, could improve the performance of the existing QA models.
Surprisingly, when using the additional evidence passages, the improvements of all the existing multi-hop reading approaches are rather limited, with the highest error reduction of 5.8\% on F1 (corresponding to 1.3\% absolute improvement) from the BERT model. 

To better understand whether the reasoning chains could indeed  help find  correct answers, we further develop a co-matching-based method that leads to 13.1\% error reduction with passage chains when applied to two of our base readers (including BERT). Our results demonstrate the existence of the potential improvement using explicit multi-hop reasoning and the necessity to develop models with better reasoning abilities.\footnote{Code and data released at \url{https://github.com/helloeve/bert-co-matching}.}

\end{abstract}

\section{Introduction}
More recent development of QA systems \cite{song2018exploring, de2018question, zhong2019coarse} has started to focus on multi-hop reasoning on text passages, aiming to propose more sophisticated models beyond the shallow matching between questions and answers. Multi-hop reasoning requires the ability to gather information from multiple different passages to correctly answer the question, and generally the task would be unsolvable by using only similarities between the question and answer.
Recent multi-hop QA datasets, such as WikiHop~\cite{welbl2018constructing}, ComplexWebQuestions~\cite{talmor2018repartitioning}, and HotpotQA \cite{yang2018hotpotqa}, have accelerated the rapid progress of QA models for multi-hop reasoning problems.

There have been several reading comprehension models proposed to address the problem. Some methods \cite{yang2018hotpotqa,zhong2019coarse} rely on cross-attention among the question and evidence passages. BERT \cite{devlin2018bert} is one successful model of such an approach.
Moreover, a substantial amount of query reformulation approaches \cite{weston2014memory,wu2016ask, shen2017reasonet,das2018multistep} have been proposed. Most of these methods adopt a soft version of reformulation, i.e. modifying the question embeddings based on the attention computed from each reasoning step. Similarly, some hard query reformulation approaches \cite{buck2018ask}  propose to rewrite the question in the original language space. These methods provide more transparency to the reasoning processes. However, their performance usually lags behind their soft counterparts when no supervision on re-writing is available.

This paper aims to investigate the following two questions for multi-hop reasoning QA systems:

\textbf{\emph{Do existing models indeed have the multi-hop reasoning ability?}}
To answer this question, we design a dataset with chains of passages ordered by the ground-truth reasoning path. Then we conduct the comparisons between two settings: (1) training and evaluating the models with the correct ordering of the passage chains~(\textbf{the ordered-oracle setting}); (2) training and evaluating the models with only the single passage that contain the answer~(\textbf{the single-oracle setting}).
We hypothesize that if the dataset indeed requires multi-hop reasoning and if a model could conduct multi-hop reasoning, it should perform significantly better in the first setting. 
However, we discovered that, for all the existing multi-hop reading comprehension models, the performance improvement with the ordered passages is rather limited, with the highest F1 improvement from BERT as 1.29\%.

\textbf{\emph{Is it beneficial to explore the usage of the reasoning chains?}}
To answer this question, we try to find a reader model which could indeed make a better use of the the ordered passage information to improve performance.
Inspired by the recent progress on the co-matching approaches for answer option selection \cite{wang2018co, zhang2019dual}, we propose to adopt a similar idea for multi-hop question answering. We extend both the HotpotReader \cite{yang2018hotpotqa} and the BERT model~\cite{devlin2018bert} with co-matching and observe 3.88\% and 2.91\% F1 improvement in the ordered-oracle setting over the single-oracle setting.
These results confirm that the utilization of passage chains is important for multi-hop question answering, and there is  untapped potential of designing new models that could perform ``real'' multi-hop reasoning.

\section{Analysis Methods} \label{analysis-method}
The goal of this analysis is to validate each model's multi-hop reasoning ability by a specifically designed dataset with three comprehensive experiment settings.

\subsection{Dataset}

We conduct the analysis over a recently released multihop QA dataset HotpotQA~\cite{yang2018hotpotqa}. 
We created a new empirical setting based on the HotpotQA distractor setting: for each question-answer pair, two supporting passage are labeled by human annotators that are sufficient for answering the question. 
We release the data of our analysis setting, to make our results comparable for future works.\footnote{\url{https://gofile.io/?c=FDsda1}.}

There have been several multi-hop QA datasets released, but none of them has the ground truth reasoning chains annotated. The reason we choose HotpotQA is that the provided supporting passages serve as a good start point for identifying the approximately correct reasoning chain of passages, based on the heuristics described below.\footnote{The HotpotQA also contains a subset of \emph{comparison} questions, which aims to select between two options by comparing a property of theirs queried by the question, e.g., \emph{Did LostAlone and Guster have the same number of members?}. These questions are not typical multi-hop questions by our community from the view of deduction. Therefore in this analysis we focus on non-comparison questions.}

The key idea to recover the reasoning chain is that the chain must end at a passage that contains the answer.
Specifically, given a question-answer pair $(q, a)$ and its two supporting passages\footnote{This heuristic only works for chains of length 2. To investigate longer chains, more complex rules are required to deal with noise in distant supervision. Popular datasets generally do not require more than 2 hops to answer questions correctly. For example all the questions in HotpotQA has no more than 2 hops. We thus leave this to future work.} $p_0$, $p_1$. Each passage $p_i$ is an abstract paragraph of a Wikipedia page, thus corresponding to a topic entity $e_i$ that is the title of the page. 
To determine the reasoning chain of passages, we have the following steps:

\noindent$\bullet$ We first check whether the answer $a$ appears in any of the passages. If there is only one passage $p_i$ containing the answer, then we have a reasoning chain with  $p_i$ as the final link of the chain, i.e., $p_{1-i} \rightarrow p_i$.

\noindent$\bullet$ If both passages contain $a$, then we use the following rule to determine the order: we check whether topic entity $e_i$ appears in $p_{1-i}$. If true, we have the chain $p_{1-i} \rightarrow p_i$. If there are still multiple matches, we simply discard the question.

For a chain $p_i \rightarrow p_j$, we denote the first passage as the \textbf{context passage} and the second as the \textbf{answer passage}.

\subsection{Analytical Method for the Ability of Multi-Hop Reasoning}
\label{ssec:analysis_method}
Based on the aforementioned dataset, we propose a systematical approach to assess the multi-hop reasoning ability of different QA models. We design three experiment settings for different passage chain compositions.

\noindent $\bullet$ \textbf{Single-Oracle}, similar to the conventional QA setting that only the question and answer passage are provided while any context passages are omitted.

\noindent $\bullet$ \textbf{Ordered-Oracle}, that the question and the extracted ordered context and answer passages are provided. 

\noindent $\bullet$ \textbf{Random}, similar to \textbf{Ordered-Oracle} but the passages are randomly ordered.

Based on the three settings,\footnote{Please note that both the Single-Oracle and the Ordered-Oracle settings are not valid realizations of the full task since they require a-priori knowledge of the answers. The settings are used in this paper only for analysis purpose.} we conduct the following analysis that each answers a research question related the multi-hop ability of the reading comprehension models:

First, we evaluate existing models on these settings, to answer the question \textbf{\emph{Q1: whether the existing models have the multi-hop reasoning ability}}. To answer the question, we mainly look at the gap between \emph{Single-Oracle} and \emph{Ordered-Oracle}.
A model with strong multi-hop reasoning capacity should have better performance in the \emph{Ordered-Oracle} setting as the reasoning path is given.

Second, if the existing methods do not show great improvement when the reasoning paths are given, we will hope to confirm \textbf{\emph{Q2: whether our dataset does not require multi-hop reasoning because of some data biases}} (see Section \ref{sec:discussion} for examples and discussions of the biases).
It is difficult to directly answer Q2, therefore in our analysis we try to answer a relevant question
\textbf{\emph{Q2$'$: whether the existing models can be further improved on the same dataset with better reasoning techniques}}.
Obviously, if there exists a technique that does better with the oracle-order information. 
It shows the reasoning paths can indeed introduce additional information in our settings, therefore the answer to \emph{Q2} is likely \emph{yes}.
Therefore our dataset and settings can be used as a criterion for evaluating different models' multi-hop reasoning ability, i.e. used for answering \emph{Q1}.

\section{Baseline Models}

For all methods, there are three inputs for the model: $q$ represents the question, $p_1$ the context passage, and $p_2$ the answer passage. Accordingly, the word-level encoded hidden sequences for these three inputs are $H^{q} \in \mathbb{R}^{l \times Q}$, $H^{p_1} \in \mathbb{R}^{l \times P_1}$, and $H^{p_2} \in \mathbb{R}^{l \times P_2}$ respectively.

\subsection{Baseline Models}
\paragraph{Bi-Attention Reader (HotpotReader)} One common state-of-the-art QA system is the HotpotReader \cite{yang2018hotpotqa} which is reported to benefit from the context passages. The system includes self-attention and bi-attention which are the standard practice in many question answering systems. We take this as one baseline as many other methods \cite{liu2017stochastic, xiong2017dcn+} generally have similar model architectures.

\paragraph{BERT Reader} Another strong baseline is to use the pre-trained BERT model to encode $q$, $p_1$, and $p_2$ all together, expecting the inner-attention mechanism to capture the order information.

Given the fact that BERT could only take one input which contains the question and answer separated by ``[SEP]", one straightforward approach to encode all three inputs by concatenating the two passages $p_1$ and $p_2$ to form the answer text ``$q$ [SEP] $p_1$ $p_2$". A more explicit way to introduce the separation of the two passages is to include a learnable boundary token by using the reserved token ``[unused0]". Therefore we design another input for BERT as ``$q$ [SEP] $p_1$ [unused0] $p_2$". We adopt both  approaches for completeness.

\section{Multi-hop Reasoning Approaches}
We seek to extend these two baseline models with two commonly used approaches for multi-hop reasoning, i.e.
query-reformulation and co-matching.

\subsection{Query-Reformulation Approach}

Query-reformulation is an idea widely used in many multi-step reasoning QA models~\cite{wu2016ask,shen2017reasonet,das2018multistep}.
The key idea is that after the model reads a paragraph, the question representation should be modified according to the matching results between the question and the paragraph. In this way, when the next paragraph comes, the model could focus on ``what is not covered'' from the history.

Most of the previous methods represent the question as a single vector so that the reformulation is performed in the embedding space. 
However, representing a question with a single vector performs badly in our task, which is not surprising since most of the top systems on recent QA leaderboards adopt word-by-word attention mechanisms.

Therefore, to have a fair comparison, we need to extend the existing methods from reformulating single vectors to reformulating the whole hidden state sequences $H^q$.
To compare the first passage $H^{p_1}$ with the question $H^q$, we applied the $BiAtt$ function and result in the matching states $\tilde{H}^q \in \mathbb{R}^{l\times Q}$, where each $\tilde{H}^q[:,i]$ states how the $i$th word of the question is matched by the passage $p_1$. Then we use these matching states to reformulate the $H^q$ as follows:
\begin{equation}
\small
    \begin{aligned}
    \tilde{H}^{q} &= BiAtt(H^{p_1}, H^q)\\
    M^q &=\gamma H^q + (1-\gamma) \mathrm{tanh}(W[H^q:\tilde{H}^{q}:H^q-\tilde{H}^{q}]) \\
    \tilde{H}^{p_2} &= BiAtt(M^{q}, H^{p_2})\\
    M &= BiLSTM(\tilde{H}^{p_2})\\
    M' &= SelfAtt(M)
    \end{aligned}
\label{eq:soft_reform}
\end{equation}
where $\gamma = \sigma(W_g[\tilde{H}^{q}:{H}^{q}:H^q-\tilde{H}^{q}])$ is a gate function. For the reformulation equation of $M^q$, we have also tried some other popular options, including only with $M^q = \mathrm{tanh}(W[H^q:\tilde{H}^{q}:H^q-\tilde{H}^{q}])$, $M^q=BiLSTM[\tilde{H}^{q}:{H}^{q}:H^q-\tilde{H}^{q}]$ and directly set
$M^q = \tilde{H}^{q}$. Among them, our gated function achieves the best performance.

\subsection{Co-Matching Approach}
The work from \cite{wang2018co} proposed a co-matching mechanism which is used to jointly encode the question and answer with the context passage. We extend the idea to conduct the multi-hop reasoning in our setup. Specifically, we integrate the co-matching to the baseline readers by firstly applying bi-attention described in Equation \ref{bi-attention} on  ($H^{q}$, $H^{p_2}$), and ($H^{p_1}$, $H^{p_2}$) using the same set of parameters.
\begin{equation}
\small
    \begin{aligned}
      \bar{H}^{q} &= {H}^{q}{G}^{q} \\
      {G}^{q} &= SoftMax(({W}^{g}{H}^{q} + {b}^{g}\otimes{e}_{p_2})^T{H}^{p_2}) \\
      \bar{H}^{p_1} &= {H}^{p_1}{G}^{p_1} \\
      {G}^{p_1} &= SoftMax(({W}^{g}{H}^{p_1} + {b}^{g}\otimes{e}_{p_2})^T{H}^{p_2}) 
    \end{aligned}
\label{bi-attention}
\end{equation}
where ${W}^{g} \in \mathbb{R}^{l \times l}$ and ${b}^{g} \in \mathbb{R}^{l}$ are learnable parameters and ${e}_{p_2} \in \mathbb{R}^{P_2}$ denotes a vector of all $1$s and it is used to repeat the bias vector into the matrix.

We further concatenate the two output hidden sequences $\bar{H}^{q}$ and $\bar{H}^{p_1}$, followed by a BiLSTM model to get the final hidden sequence for answer prediction as shown in Equation \ref{co-match}. The start and end of the answer span is predicted based on $M$.
\begin{equation}
\small
    M = BiLSTM([\bar{H}^{q}:\bar{H}^{p_1}]) \\
\label{co-match}
\end{equation}

\paragraph{Co-Matching in HotpotReader}
We follow the above co-matching approach on the HotporReader's output directly.

\paragraph{Co-Matching in BERT}
One straightforward way to achieve co-matching in BERT is to separately encode the question, the first passage and the second one with BERT, and then apply the above co-matching functions on the output hidden sequence as proposed in \cite{zhang2019dual}.

However, as observed in the experiments, we believe the inter-attention mechanism (i.e. cross paragraph attention) could capture the order information in an implicit way. Therefore, we still hope to benefit from the cross passage attention inside BERT, but make it better cooperate with three inputs. After the original encoding from BERT, we apply the co-matching\footnote{To follow the original BERT's setup, we also apply the same attention dropout with a probability of 0.9 on the attention scores.} on the output sequence to explicitly encourage the reasoning path. $H^{q}$, $H^{p_1}$, and $H^{p_2}$ could be easily obtained by masking the output sequence according to the original text.

\section{Experiments}
\subsection{Settings}
We trained and evaluated each model for comparison for each setting separately.
Following previous work \cite{yang2018hotpotqa}, we report the exact-match and F1 score for the answer prediction task.

\subsection{Results}
\label{ssec:exp_results}
In Table \ref{baseline-result}, 
the original HotpotReader method does not show significant performance improvement when comparing the Single-Oracle setting with the Ordered-Oracle setting. BERT was able to get a small improvement from its inner cross passage attention which introduces some weak reasoning. Surprisingly, overall the context passage in the reasoning path  does not inherently contribute to the performance of these methods, which indicates that the models are not learning much multi-hop reasoning as previously thought.

\begin{table}[!htbp]
\small
\centering
\begin{tabular}{lcccc}
\toprule
\multirow{2}{*}{\bf Model} &  \multicolumn{2}{c}{\bf Single-Oracle} & \multicolumn{2}{c}{\bf Ordered-Oracle} \\
& \bf EM & \bf F1 & \bf EM & \bf F1 \\
\midrule
HotpotReader & 55.07 & 70.00 & 55.17 & 70.75 \\
Bert & 64.08 & 77.86 & 65.03 & 79.15 \\
\bottomrule
\end{tabular}
\caption{Baseline results for HotpotReader and BERT}
\label{baseline-result}
\end{table}

We show our proposed improvements in Table \ref{hotpotreader-result} and \ref{bert-result}. 
Compared to the Single-Oracle baseline (HotpotReader), when applying the co-matching mechanism in
the Ordered-Oracle setting, there is a significant improvement of 4.38\% in exact match and 4.26\% in F1.
The soft query reformulation also improves the performance but not as significantly.
In order to confirm that the improvement of co-matching does come from the usage of reasoning paths (instead of the higher model capacity), we make another comparison that runs the co-matching model over the Single-Oracle setting. To achieve this, we duplicate the single oracle passage twice as $p_1$ and $p_2$. Our results show that this method does not give any improvement.
Therefore the co-matching method indeed contributes to the performance gain of multi-hop reasoning.

\begin{table}[!htbp]
\small
\centering
\begin{tabular}{lcccc}
\toprule
\multirow{2}{*}{\bf Model} & \bf Order &\multicolumn{2}{c}{\bf Performance}\\
    & & \bf EM  & \bf F1  \\
\midrule
\multirow{3}{*}{HotpotReader} & Random  & 52.23 & 69.80 \\
& Single-Oracle & 55.07 & 70.00\\
& Ordered-Oracle & 55.17 & 70.75\\
\midrule
\quad w/ Query-Reform & Ordered-Oracle & 56.89 & 71.69 \\
\midrule
\multirow{2}{*}{\quad w/ Co-Matching} & Single-Oracle & 55.00 & 70.23 \\
& Ordered-Oracle & \bf 59.45 & \bf 74.26 \\
\bottomrule
\end{tabular}
\caption{Results for HotpotReader on 3 oracle settings}
\label{hotpotreader-result}
\end{table}

BERT achieved promising results even in the Single-Oracle setting which proves its original capacity for QA. 
The original BERT was improved by 1.23\% in exact match when both context passage and answer passage are provided and separated by an extra token. Nonetheless,
the co-matching mechanism contributes to an additional 1.66\% exact match improvement which indicates the success of co-matching for reasoning. Co-matching result also shows that multi-hop over passage chain contains additional information, and thus multi-hop ability is necessary in our analysis setting.

\begin{table}[!htbp]
\small
\centering
\begin{tabular}{lcccc}
\toprule
\multirow{2}{*}{\bf Model} & \bf Order &\multicolumn{2}{c}{\bf Performance}\\
    & & \bf EM  & \bf F1  \\
\midrule
\multirow{2}{*}{BERT} & Random  & 59.18 & 75.27 \\
& Single-Oracle & 64.08 & 77.86\\
& Ordered-Oracle & 65.03 & 79.15\\
\quad w/ split token & Ordered-Oracle & 65.31 & 79.49 \\
\midrule
\quad w/ Co-Matching & Ordered-Oracle & \bf 66.97 & \bf 80.77 \\
\bottomrule
\end{tabular}
\caption{Results for BERT on 3 oracle settings}
\label{bert-result}
\end{table}

Among both approaches, co-matching shows promising performance improvement especially for the well pre-trained BERT model. This proves the co-matching mechanism is able to conduct multi-hop reasoning following the passage chains.

Finally, both models perform worse in the Random setting compared to the Single-Oracle setting, although the Random setting contains sufficient information of the whole reasoning chain. From the analysis, we find that it is difficult for the models to correctly predict the orders from the randomly-ordered passages. For example, we created a binary classification task to predict which passage is the context passage and which is the answer passage. BERT model gives an accuracy of only 87.43\% on this task.
This gives further evidence that the existing models do not have appropriate inductive biases for utilizing the reasoning chains.

\paragraph{Answers to our research questions}
The above results answer our research questions as follows: (1) in our experimental setting, the reasoning paths are indeed useful, thus multi-hop reasoning is necessary, as there exists a method, i.e., co-matching, that has demonstrated significant improvement; (2) existing reader models usually cannot fully make use of the reasoning paths, indicating their limited reasoning abilities. Among the existing methods, BERT can do slightly better on making use of the reasoning paths. Our new proposed co-matching approach improves the reasoning abilities over both the two different base models (HotpotReader and BERT).

\section{Discussion}
\label{sec:discussion}

\paragraph{Difference from prior work}
Our work conducts the first analysis of \emph{models' behaviors}.
In comparison, a concurrent analysis work~\cite{min2019compositional}, which is also conducted on HotpotQA, focuses more on the properties of the dataset.
For example, \citep{min2019compositional} finds that for 80\% of the questions in HotpotQA, humans do not need the full paths of paragraphs to correctly answer some of the questions.
One of the major reasons is the bias of factoid questions that look for certain types of entities as answers. For example, a question asking ``\emph{which sports team}'' can be directly answered if there is only one sports team mentioned in the documents.

Our analysis focuses on whether the full reasoning paths can help the \emph{machine learning models} to (1) improve their performance on those 80\% of the questions, as well as (2) cover the left 20\% of questions that indeed require the multi-hop ability.
Moreover, compared to the prior analysis, we are the first to analyze the effects of reasoning paths in an explicit way, and construct a dataset for this purpose.

\paragraph{The effect of data biases on our analysis}

The aforementioned biases make the full reasoning paths less useful for a large portion of data, therefore making it more challenging for reader models to improve with full reasoning paths.

Because of the data bias, it is critical to verify that the dataset we created can still benefit from the improved reasoning skills. That is why answering \emph{Q2} in Section \ref{ssec:analysis_method} is important for the whole analysis.
The results in Section \ref{ssec:exp_results} show that our co-matching methods can indeed benefit from the reasoning paths, confirming the effectiveness of our proposed dataset and settings for the analysis purpose.

\paragraph{Encouraging model design with better evaluation}
Finally, continued from the previous paragraph, we hope to highlight the problem that the less biased a dataset is, the more likely a model can easily benefit from the availability of reasoning paths.
On many existing benchmark datasets that are biased, it is less likely to achieve improvement with specific designs for achieving multi-hop reasoning ability.
This makes multi-hop reasoning a less important factor when people design models for these multi-hop QA datasets, if the goal is simply to improve the answer accuracy.

To encourage model design towards real reasoning instead of fitting the data biases, we believe that an improved evaluation is necessary. To this end, one way is certainly to create datasets with fewer biases. While our analysis also suggests the other way: we can keep the biased training data, but created small evaluation datasets with human-labeled reasoning paths. Then during evaluation, we compute the accuracy of the predicted reasoning paths. This is an extension of the idea of HotpotQA that jointly evaluates the support selection and answer extraction, but with a more explicit focus on the reasoning processes.

\section{Conclusion}
In this paper, we analyze QA models' capability in multi-hop reasoning by assessing if the reasoning chain could help existing multi-hop readers. We observed the general weakness of stat-or-the-art models in multi-hop reasoning and proposed a co-matching based method to mitigate. Despite the fact that co-matching is designed to encode only three input sequences to achieve limited multi-hop reasoning, we consider this as the most promising one that demonstrates the concrete reasoning capability and has the potential for real multi-hop reasoning. 

\section*{Acknowledgments}
We thank the anonymous reviewers for their very valuable comments and suggestions.

\bibliography{emnlp-ijcnlp-2019}

\begin{thebibliography}{17}
\expandafter\ifx\csname natexlab\endcsname\relax\def\natexlab#1{#1}\fi

\bibitem[{Buck et~al.(2018)Buck, Bulian, Ciaramita, Gajewski, Gesmundo,
  Houlsby, and Wang.}]{buck2018ask}
Christian Buck, Jannis Bulian, Massimiliano Ciaramita, Wojciech Gajewski,
  Andrea Gesmundo, Neil Houlsby, and Wei Wang. 2018.
\newblock Ask the right questions: Active question reformulation with
  reinforcement learning.
\newblock In \emph{International Conference on Learning Representations}.

\bibitem[{Das et~al.(2019)Das, Dhuliawala, Zaheer, and
  McCallum}]{das2018multistep}
Rajarshi Das, Shehzaad Dhuliawala, Manzil Zaheer, and Andrew McCallum. 2019.
\newblock Multi-step retriever-reader interaction for scalable open-domain
  question answering.
\newblock In \emph{International Conference on Learning Representations}.

\bibitem[{De~Cao et~al.(2018)De~Cao, Aziz, and Titov}]{de2018question}
Nicola De~Cao, Wilker Aziz, and Ivan Titov. 2018.
\newblock Question answering by reasoning across documents with graph
  convolutional networks.
\newblock \emph{arXiv preprint arXiv:1808.09920}.

\bibitem[{Devlin et~al.(2018)Devlin, Chang, Lee, and
  Toutanova}]{devlin2018bert}
Jacob Devlin, Ming-Wei Chang, Kenton Lee, and Kristina Toutanova. 2018.
\newblock Bert: Pre-training of deep bidirectional transformers for language
  understanding.
\newblock \emph{arXiv preprint arXiv:1810.04805}.

\bibitem[{Liu et~al.(2017)Liu, Shen, Duh, and Gao}]{liu2017stochastic}
Xiaodong Liu, Yelong Shen, Kevin Duh, and Jianfeng Gao. 2017.
\newblock Stochastic answer networks for machine reading comprehension.
\newblock \emph{arXiv preprint arXiv:1712.03556}.

\bibitem[{Min et~al.(2019)Min, Wallace, Singh, Gardner, Hajishirzi, and
  Zettlemoyer}]{min2019compositional}
Sewon Min, Eric Wallace, Sameer Singh, Matt Gardner, Hannaneh Hajishirzi, and
  Luke Zettlemoyer. 2019.
\newblock Compositional questions do not necessitate multi-hop reasoning.
\newblock \emph{arXiv preprint arXiv:1906.02900}.

\bibitem[{Shen et~al.(2017)Shen, Huang, Gao, and Chen}]{shen2017reasonet}
Yelong Shen, Po-Sen Huang, Jianfeng Gao, and Weizhu Chen. 2017.
\newblock Reasonet: Learning to stop reading in machine comprehension.
\newblock In \emph{Proceedings of the 23rd ACM SIGKDD International Conference
  on Knowledge Discovery and Data Mining}, pages 1047--1055. ACM.

\bibitem[{Song et~al.(2018)Song, Wang, Yu, Zhang, Florian, and
  Gildea}]{song2018exploring}
Linfeng Song, Zhiguo Wang, Mo~Yu, Yue Zhang, Radu Florian, and Daniel Gildea.
  2018.
\newblock Exploring graph-structured passage representation for multi-hop
  reading comprehension with graph neural networks.
\newblock \emph{arXiv preprint arXiv:1809.02040}.

\bibitem[{Talmor and Berant(2018)}]{talmor2018repartitioning}
Alon Talmor and Jonathan Berant. 2018.
\newblock Repartitioning of the complexwebquestions dataset.
\newblock \emph{arXiv preprint arXiv:1807.09623}.

\bibitem[{Wang et~al.(2018)Wang, Yu, Chang, and Jiang}]{wang2018co}
Shuohang Wang, Mo~Yu, Shiyu Chang, and Jing Jiang. 2018.
\newblock A co-matching model for multi-choice reading comprehension.
\newblock \emph{arXiv preprint arXiv:1806.04068}.

\bibitem[{Welbl et~al.(2018)Welbl, Stenetorp, and
  Riedel}]{welbl2018constructing}
Johannes Welbl, Pontus Stenetorp, and Sebastian Riedel. 2018.
\newblock Constructing datasets for multi-hop reading comprehension across
  documents.
\newblock \emph{Transactions of the Association of Computational Linguistics},
  6:287--302.

\bibitem[{Weston et~al.(2014)Weston, Chopra, and Bordes}]{weston2014memory}
Jason Weston, Sumit Chopra, and Antoine Bordes. 2014.
\newblock Memory networks.
\newblock \emph{arXiv preprint arXiv:1410.3916}.

\bibitem[{Wu et~al.(2016)Wu, Wang, Shen, Dick, and van~den Hengel}]{wu2016ask}
Qi~Wu, Peng Wang, Chunhua Shen, Anthony Dick, and Anton van~den Hengel. 2016.
\newblock Ask me anything: Free-form visual question answering based on
  knowledge from external sources.
\newblock In \emph{Proceedings of the IEEE Conference on Computer Vision and
  Pattern Recognition}, pages 4622--4630.

\bibitem[{Xiong et~al.(2017)Xiong, Zhong, and Socher}]{xiong2017dcn+}
Caiming Xiong, Victor Zhong, and Richard Socher. 2017.
\newblock Dcn+: Mixed objective and deep residual coattention for question
  answering.
\newblock \emph{arXiv preprint arXiv:1711.00106}.

\bibitem[{Yang et~al.(2018)Yang, Qi, Zhang, Bengio, Cohen, Salakhutdinov, and
  Manning}]{yang2018hotpotqa}
Zhilin Yang, Peng Qi, Saizheng Zhang, Yoshua Bengio, William~W Cohen, Ruslan
  Salakhutdinov, and Christopher~D Manning. 2018.
\newblock Hotpotqa: A dataset for diverse, explainable multi-hop question
  answering.
\newblock \emph{arXiv preprint arXiv:1809.09600}.

\bibitem[{Zhang et~al.(2019)Zhang, Zhao, Wu, Zhang, Zhou, and
  Zhou}]{zhang2019dual}
Shuailiang Zhang, Hai Zhao, Yuwei Wu, Zhuosheng Zhang, Xi~Zhou, and Xiang Zhou.
  2019.
\newblock Dual co-matching network for multi-choice reading comprehension.
\newblock \emph{arXiv preprint arXiv:1901.09381}.

\bibitem[{Zhong et~al.(2019)Zhong, Xiong, Keskar, and Socher}]{zhong2019coarse}
Victor Zhong, Caiming Xiong, Nitish~Shirish Keskar, and Richard Socher. 2019.
\newblock Coarse-grain fine-grain coattention network for multi-evidence
  question answering.
\newblock \emph{arXiv preprint arXiv:1901.00603}.

\end{thebibliography}
\bibliographystyle{acl_natbib}

\appendix

\end{document}